\documentclass[11pt]{article}

\usepackage[preprint]{acl}

\usepackage{times}
\usepackage{latexsym}
\usepackage[T1]{fontenc}
\usepackage[utf8]{inputenc}
\usepackage{microtype}
\usepackage{inconsolata}
\usepackage{graphicx}

\usepackage{amsmath}
\usepackage{caption}
\usepackage{subcaption}
\usepackage{multirow}
\usepackage{enumitem}
\usepackage{listings}
\usepackage{xcolor}
\usepackage{booktabs}

\setlist[itemize]{leftmargin=1.5em, topsep=0.5em, itemsep=0em}

\newcommand{\tighttt}[2][8pt]{
  {\fontsize{#1}{#1}\selectfont
   \ttfamily
   \spaceskip=0.3em\relax
   \kern-0.2em#2\kern-0.2em
  }
}

\title{Towards Autopilot Labeling for Short Video Content Governance}

\author{Junwen Chen\thanks{Equal contribution.}, Hongwei Wang\footnotemark[1], Zixuan Wang, Jingbing Wang, Gorden Liu,\\{\bf Sheng Siong Ong, Xiaojiao Zang, JuChin Chao, Ze Jin, Kaiwen Men, Yin Li},\\{\bf Jeremy Li, Zhuolin Hao, Hongyu Xiong, Yang Song} \\
  TikTok Inc.\\
  San Jose, CA, USA \\
  \texttt{\{chenjunwen.javen, hongwei.w, hongyu.xiong\}@tiktok.com}}

\begin{document}
\maketitle
\begin{abstract}
    The explosive growth of short videos has presented significant challenges to content governance.
    Traditional manual labeling is inefficient, slow to iterate, and costly.
    Therefore, we explore and develop a multi-stage \textit{autopilot labeling} framework aimed at progressively increasing the level of automation in short video content annotation.
    To move beyond the heavy reliance on human expert labeling at the initial \textit{L0 stage}, we first introduce a discriminative MLLM labeler at the \textit{L1 stage}, which is fine-tuned with supervised data to directly predict labels for short videos.
    Advancing to the \textit{L2 stage}, we shift to a domain-adaptive MLLM pretrained on short video data and use it for zero-shot labeling, enabling flexible generalization without task-specific supervision.
    In the \textit{L3 stage}, we propose an agentic annotation workflow, and equip the agent with several key tools, which empower the agent to handle complex, unfamiliar, or vaguely defined cases, significantly reducing the need for human intervention.
    We conduct extensive experiments across various content issues, and results demonstrate that each stage of the autopilot labeling framework progressively improves labeling accuracy, and has significantly reduced manual labor costs in real-world applications by up to 85\%.
\end{abstract}

\begin{figure*}[t]
    \centering
    \includegraphics[width=1\textwidth]{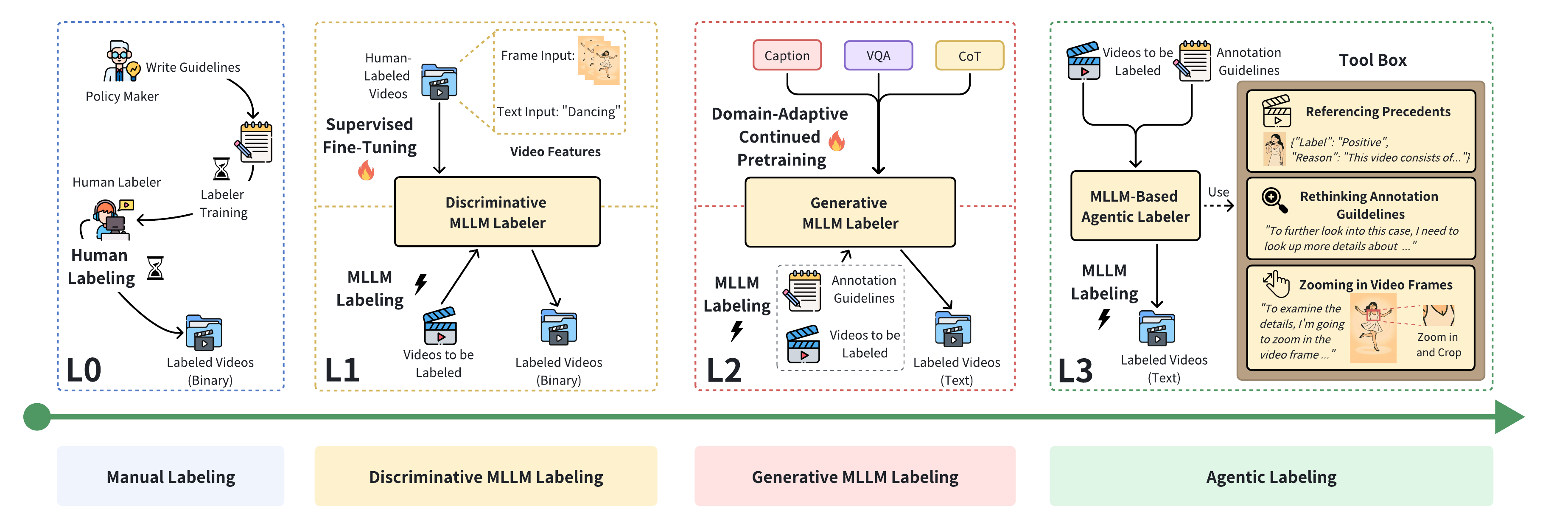}
    \vspace{-2em}
    \caption{Illustration of our multi-stage autopilot labeling framework for short video content governance. \includegraphics[height=0.7em]{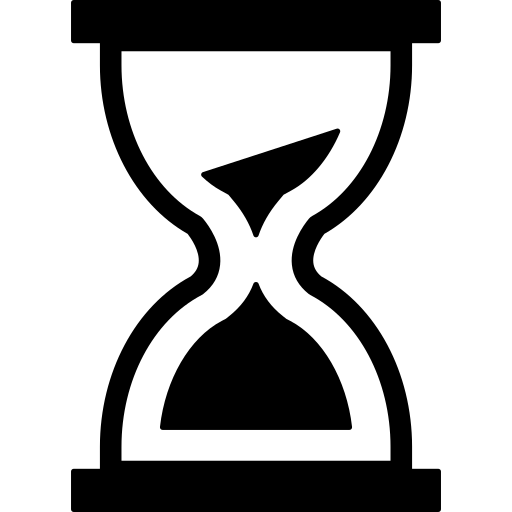} indicates a longer processing time, while \includegraphics[height=0.7em]{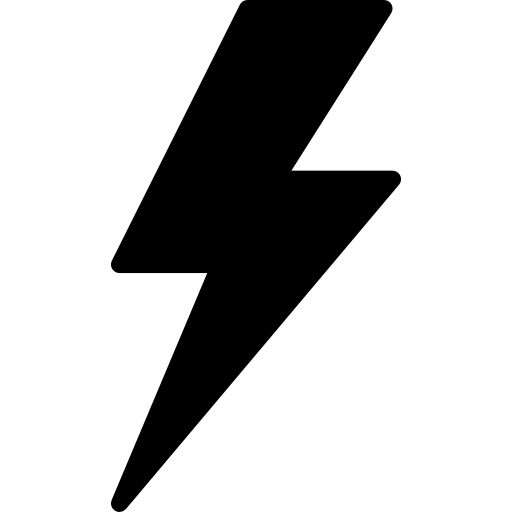} indicates a shorter processing time.}
    \label{fig:main}
\end{figure*}

\section{Introduction}
    
    Users on short video platforms upload a large number of videos every day.
    Some of these videos may violate platforms' guidelines for maintaining a healthy platform ecosystem, such as sexually suggestive content, non-original material, clickbait, and poor-quality content.
    Videos identified as violations are subject to reduced visibility and, in some cases, may be removed from the platform.

    Currently, we are entirely relying on human annotators to monitor the density of problematic videos online, and to prepare labels to train downstream models for issue governance.
    However, manual annotation comes with several drawbacks, such as low accuracy and consistency, inefficiency, high costs, and long iteration cycles.
    With the rapid development of multimodal large language models (MLLMs) \citep{achiam2023gpt, liu2023visual, wang2024qwen2, team2023gemini,sun2025audio,jing2026radar}, the capability for video content understanding has significantly improved, making automated video annotation increasingly feasible.
    This paper aims to introduce our practice in exploring automated annotation for short videos.

    As illustrated in Figure \ref{fig:main}, similar to the different levels in autonomous driving, our approach to short video annotation has also advanced step by step.
    The \textit{L0 stage} represents no automation, where all labeling is done entirely by humans.
    
    The \textit{L1 stage} introduces limited model assistance to support human annotators, which marks our first step toward automated labeling.
    Specifically, we train discriminative MLLMs on supervised data to directly predict labels.
    Compared to online governance models, these MLLM labelers are much larger in size and run offline to generate training data for the online models.

    However, L1-stage models still suffer from limited accuracy and remain heavily dependent on a constant supply of human-labeled data, which restricts its level of automation.
    In the \textit{L2 stage}, we shift to the generative use of MLLMs, enabling them to produce natural language labels and explanations.
    This improves the automation of the labeling process, as we no longer need continuous manual annotations to train labeling models.
    We conduct domain-adaptive continued pretraining on native MLLMs, then perform zero-shot labeling.

    While the zero-shot approach in the L2 stage performs well on simple cases, it still struggles with complex, unseen, or ambiguously defined ones.
    To address this, we introduce an LLM-based agent workflow in the \textit{L3 stage}.
    To make the agentic labeler more powerful and flexible, we design three useful tools: \textit{referencing precedents (examples)}, \textit{rethinking annotation guidelines}, and \textit{zooming in video frames}.
    This agent design significantly enhances the automation of annotation process.

    We conduct experiments on four content issues.
    The results show that the performance of our labeling models improves progressively across different stages of the autopilot framework.
    Furthermore, our results indicate that, when maintaining the same labeling quality as human annotators, incorporating model-generated labels can reduce human labor costs by 27\% to 85\%.
    We have already deployed the labeling system in the production environment, where human-MLLM collaboration has significantly reduced labor costs.

\section{Related Work}

    \vspace{1ex}
    \noindent \textbf{Multi-Modal LLMs}.
    MLLMs have rapidly advanced in aligning and generating across vision and language modalities.
    Contrastive models like CLIP \citep{radford2021learning} lay the foundation by aligning image-text pairs at scale.
    Subsequent models such as Flamingo \citep{alayrac2022flamingo} and BLIP \citep{li2022blip} incorporate generative capabilities and few-shot reasoning.
    LLaVA \citep{liu2023visual} and Qwen-VL series \citep{bai2023qwen, wang2024qwen2, liu2023visual} enhance multimodal instruction following by combining visual encoders with large language models.
    GPT-4V \citep{achiam2023gpt} pushes further toward general-purpose agents capable of interleaved multimodal reasoning.
    These MLLMs can serve as base models for our continued pretraining and be used as our labelers.
    
    \vspace{1ex}
    \noindent \textbf{LLM-Based Agents}.
    LLM-based agents have gained significant attention for their ability to reason, plan, and interact within diverse environments \citep{wang2025comprehensive,zhang2025cognitive}.
    Early frameworks like ReAct \citep{yao2023react} introduce a framework combining reasoning traces with actions, significantly improving task performance.
    Toolformer \citep{schick2023toolformer} enables tool use through reasoning traces and self-supervised learning.
    Approaches like Voyager \citep{wang2023voyager} apply these ideas in interactive environments.
    AgentBench \citep{liu2023agentbench} conducts environments to assess LLM-as-Agent’s reasoning and decision-making abilities in a multi-turn openended generation setting.
    Furthermore, advancements in memory-augmented agents \citep{zhang2024cognitive} have enhanced their adaptability and long-term planning.
    We believe that the agent's reasoning ability and its autonomous tool invocation capability are the key factors in achieving fully autonomous short video annotation.

\section{Approaches}
\label{sec:approaches}
\subsection{Manual Labeling: L0 Stage}
    \label{sec:l0}

    There are various issues on our platform that require governance.
    Let $\mathcal I$ denote the set of all such issues.
    For each issue $i \in \mathcal I$, the governance team defines a detailed description and a corresponding annotation guideline, denoted as $G^i$.
    A human annotator $H$ is trained on the guidelines and then labels a given video $v$ with a binary label $l_{binary}$:
    \begin{equation}
        l_{binary} = H (v; G^i).
    \end{equation}
    The manual annotation workflow is introduced in Appendix \ref{app:l0}.

    \vspace{1ex}
    The limitations of manual labeling are significant: \textit{low accuracy and consistency}, \textit{slow and inefficient}, \textit{high operational cost}, and \textit{long iteration cycles}.
    The detailed discussion is in Appendix \ref{app:l0}.

\subsection{Discriminative MLLM Labeling: L1 Stage}
    \label{sec:l1}
    
    Our first step toward automated labeling is using an MLLM to directly classify an input video $v$:
    \begin{equation}
        l_{binary} = M_{dis}^i (v),
    \end{equation}
    where $M_{dis}^i$ is a discriminative MLLM labeler for issue $i$.
    $M_{dis}^i$ has a superscript $i$ since it is issue-specific.
    This approach is simple and straightforward to implement.
    Moreover, it produces structured classification outputs that can seamlessly integrate into existing human labeling workflows.

    Specifically, we use LLaVA \citep{liu2023visual} as the base model and add a classification head to predict the probabilities for each class.
    As shown in the L1 stage in Figure \ref{fig:main}, each short video input is composed of two types of features:
    \textit{textual features} (title, sticker text, OCR text, ASR transcripts) and \textit{visual features} (frames sampled from the video).

    More discussion on discriminative MLLM labelers, including their relationship with governance models and how they save manual labeling costs, is in Appendix \ref{app:l1}.

\subsection{Generative MLLM Labeling: L2 Stage}
    \label{sec:l2}

    Discriminative MLLM labeling has several limitations:
    A separate labeler must be trained for each issue, resulting in multiple models to maintain;
    Moreover, since the model still relies on human-annotated data for training, its development and iteration remain time-consuming (See Appendix \ref{app:experiment_ablation} for empirical analysis).
    
    To address these challenges, we turn to the generative use of MLLMs.
    Generative MLLMs produce labels and explanations in natural language, making them better suited for tasks where the annotation guidelines are ambiguous, diverse, or require reasoning.
    Specifically, we have
    \begin{equation}
        l_{text} = M_{gen} (v; G^i),
    \end{equation}
    where $M_{gen}$ is an all-in-one generative MLLM, $G^i$ is the annotation guideline for issue $i$, and $l_{text}$ is the generated text containing the predicted label.

    For $M_{gen}$, we do not directly use native MLLMs, because they have not been exposed to our short videos and lacks knowledge of issue definitions.
    Instead, we performed domain-adaptive continued pretraining on top of native MLLMs.
    The details are included in Appendix \ref{app:pretrain}.

    Building on our continued-pretrained MLLM, we first explore a straightforward labeling prompt:
    ``\texttt{<Annotation guidelines> <Video info> Does the input video violate any of the listed rules?}''
    However, the annotation guidelines are typically very lengthy, often consisting of dozens of items and tens of thousands of words.
    This is challenging for MLLMs, which tend to overlook important details in such long contexts.

    \vspace{1ex}
    \noindent \textbf{Annotation guidelines decomposition}.
    To mitigate this issue, we decompose the annotation guidelines into a set of smaller, more focused sub-items.
    For instance, in the case of borderline sexual content, we break it down into 17 subcategories, such as ``\textit{Adult sexualized behavior}'', ``\textit{Adult fetish \& kinks}'', and ``\textit{Adult body exposure}''.
    The MLLM generates a judgment for each sub-item with the prompt ``\texttt{<Annotation guidelines sub-item> <Video info> Does the input video violate this rule?}'', and we aggregate the results for all sub-items.
    If any single sub-item is judged as a violation, the final labeling output is marked as a positive case.
    Specifically, we aggregate the results for each sub-item $G^{i,s}$ in guideline $G^i$:
    \begin{equation}
    \begin{split}
        l_{text}^s &= M_{gen} \big( v; G^{i, s} \big),\\
        l_{binary} &= \text{any} \Big( \text{bool} \big( l_{text}^s \big) \ \text{for} \ G^{i,s} \in G^i \Big).
    \end{split}
    \end{equation}

    $G^{i,s}$ is typically under a thousand tokens, which significantly reduces the model's comprehension burden.
    Moreover, this process can be parallelized across sub-items, greatly improving the overall efficiency.

\subsection{Agentic Labeling: L3 Stage}
    \label{sec:l3}
    The zero-shot approach in the L2 stage performs well on simple cases, but struggles with complex, unseen, or ambiguously defined ones.
    Therefore, in the L3 stage, we investigate agentic work flows to handle the annotation task.
    We equip the agent with the following tools: \textit{referencing precedents}, \textit{rethinking annotation guidelines}, and \textit{zooming in video frames}.

    \vspace{1ex}
    \noindent \textbf{Referencing Precedents}.
        High-quality precedents (examples) are highly valuable for video annotation \citep{dong2022survey}.
        Zero-shot methods often struggle to identify clear decision boundaries, while textual guidelines alone can be abstract and ambiguous.
        Well-chosen examples, especially those with visual content, clearly illustrate labeling criteria and make rules easier to apply.
        Moreover, a diverse set of examples covers common scenarios, allowing the model to imitate relevant reasoning and reducing annotation difficulty.

        Specifically, for each sub-item $G^{i,s}$ in the annotation guidelines $G^i$, we now enrich the model's input with a corresponding set of examples $\{e_j\}$:
        \begin{equation}
            l^s_{text} = M_{gen} \big( v; G^{i,s}, \{e_j\} \big).
        \end{equation}

        The number of examples used is also an interesting factor that affects performance.
        Too few examples may fail to clearly define the boundary between positive and negative cases, while too many examples can burden the model with processing an overly long context.
        We will present the corresponding experimental results in Section \ref{sec:ablation}.

    \vspace{1ex}
    \noindent \textbf{Rethinking Annotation Guidelines}.
        In Section \ref{sec:l2}, we discussed how the annotation guidelines are decomposed into sub-items.
        We observe that the MLLM shows particularly low accuracy on certain sub-items, typically due to the abstract and ambiguous nature of their descriptions.
        For example, in the ``\textit{Adult body exposure}'' sub‑item, the definition for ``\textit{Light cleavage exposure}'' is ``\textit{The visibility of the inverted V, which is the area between and/or below the breasts, where cleavage is visible in low‑cut shirts, dresses, or garments.}''
        The concept of ``\textit{inverted V}'' is highly abstract, making it difficult for the model to reason properly, which leads to poor accuracy on this sub-item.

        To mitigate this issue, we introduce a ``rethinking'' tool that allows the agent to selectively revisit sub-items it may have misjudged.
        For sub-items prone to ambiguity, we provide expanded definitions and, when necessary, additional targeted examples that the agent can consult during rethinking \citep{jiang2024many}:
        \begin{equation}
        \begin{split}
            l^s_{text} =& M_{gen} \Big( v; E \big( G^{i,s} \big) \Big) \quad \text{or} \\
            \quad l^s_{text} =& M_{gen} \Big( v; E \big( G^{i,s} \big), E \big( \{e_j\} \big) \Big).
        \end{split}
        \end{equation}

        We do not apply expanded definitions by default for efficiency reasons: they are substantially longer (e.g., nearly 1k words vs. 26 words for ``\textit{inverted V}'') and include many additional examples, which would incur prohibitive computational cost.
        Instead, the agent triggers rethinking only when it lacks confidence in its initial judgment.

    \vspace{1ex}
    \noindent \textbf{Zooming in Video Frames}.
        Beyond rethinking annotation guidelines, the agent may also need to attend to fine-grained visual details in video frames.
        Each video contains 16 frames, yet evidence of a violation often appears in a small region of a single frame (e.g., the ``\textit{Inverted V}'').
        To reduce irrelevant visual context, the agent can zoom in on a localized region for focused inspection using the zoom-in tool $Z(\cdot)$:
        \begin{equation}
            l^s_{text} = M_{gen} \Big ( Z(v); G^{i,s} \Big).
        \end{equation}

        The zoom-in tool can be combined with rethinking/precedent tools for more accurate decisions:
        \begin{equation}
            l^s_{text} = M_{gen} \Big( Z(v); E \big( G^{i,s} \big), E \big( \{e_j\} \big) \Big).
        \end{equation}

        We consider two implementations of zoom-in: (1) predicting a bounding box and cropping the region \citep{zheng2025deepeyes}, or (2) selecting a cell from a fixed spatial grid (e.g., 5$\times$5).
        Since bounding boxes tend to drift and lack precision (Section \ref{sec:ablation}), we adopt the grid-based approach.

        Zoom-in is conceptually similar with rethinking: while rethinking performs a ``semantic zoom-in'' on ambiguous guideline sub-items, zoom-in performs a spatial zoom-in on localized visual evidence.

        \vspace{1ex}
        More analysis and discussion on agentic labeling is provided in Appendix \ref{app:agentic_discussion}.

    \begin{table*}[t]
        \centering
        \setlength{\tabcolsep}{4pt}
        \small
        \begin{tabular}{c|cccc|cccc|cccc}
            \toprule
            \multirow{2}{*}{\textbf{Labeling Model}} & \multicolumn{4}{c|}{\textbf{SSC}} & \multicolumn{4}{c|}{\textbf{PQC}} & \multicolumn{4}{c}{\textbf{CB}} \\
            & \textit{Acc} & \textit{P} & \textit{R} & \textit{F1} & \textit{Acc} & \textit{P} & \textit{R} & \textit{F1} & \textit{Acc} & \textit{P} & \textit{R} & \textit{F1} \\
            \midrule
            \midrule
            \makebox[17em][c]{Discriminative MLLM labeler (L1)} & 71.0 & 91.8 & 47.0 & 62.2 & \underline{93.9} & \textbf{97.4} & \underline{89.6} & \underline{93.4} & 65.6 & 59.3 & 69.9 & 64.1 \\
            \midrule
            \multicolumn{1}{c}{Generative MLLM labeler (L2)} & \multicolumn{12}{c}{} \\
            \makebox[17em][l]{\textit{Zero-shot}} & 60.2 & \textbf{96.5} & 22.1 & 36.0 & 79.1 & 86.5 & 70.0 & 75.5 & 73.2 & \textbf{70.8} & 57.6 & 63.5 \\
            \makebox[17em][l]{\quad+ \textit{Annotation guidelines decomposition}} & 71.3 & 91.8 & 47.5 & 62.6 & 83.3 & 85.6 & 78.5 & 81.9 & 72.2 & 63.6 & 72.8 & 67.9 \\
            \midrule
            \multicolumn{1}{c}{Agentic Labeler (L3)} & \multicolumn{12}{c}{} \\
            \makebox[17em][l]{\textit{Referencing precedents}} & 78.1 & 89.3 & 64.6 & 74.9 & 86.2 & 89.1 & 81.3 & 85.0 & 73.6 & 65.4 & 73.6 & 69.3 \\
            \makebox[17em][l]{\quad+ \textit{Rethinking annotation guidelines}} & 78.3 & 86.2 & \underline{68.2} & 76.1 & 86.8 & 87.2 & 85.0 & 86.1 & 73.9 & 65.2 & \underline{76.5} & 70.4 \\
            \makebox[17em][l]{\quad+ \textit{Zooming-in video frames}} & \underline{79.0} & 88.2 & 67.5 & \underline{76.5} & -  & - & - & - & \underline{74.2} & 67.5 & 76.0 & \underline{71.5} \\
            \midrule
            Combining L1 and L3 & \textbf{82.4} & 88.5 & \textbf{75.1} & \textbf{81.3} & \textbf{96.0} & \underline{96.6} & \textbf{95.0} & \textbf{95.8} & \textbf{78.4} & \underline{70.1} & \textbf{79.3} & \textbf{74.6} \\
            \midrule
            \midrule
            LLaVA-OV (L2) & 55.6 & 91.7 & 13.5 & 23.4 & 69.0 & 80.6 & 50.0 & 61.7 & 64.6 & 58.6 & 39.2 &47.0 \\
            Qwen2.5-VL (L2) & 59.1 & 94.9 & 20.2 & 33.4 & 64.6 & 70.9 & 49.6 & 58.4 & 64.7 & 60.1 & 38.0 & 46.5 \\
            GPT-5 (L2) & 73.4 & \underline{95.3} & 40.2 & 56.5 & 80.1 & 88.4 & 73.9 & 80.5 & 72.2 & 67.0 & 55.1 & 60.5 \\
            \bottomrule
        \end{tabular}
        \caption{Result of all labeling models. The evaluation metrics are Accuracy, Precision, Recall, and F1 score (in \%). The highest value in each column is shown in \textbf{bold}, and the second-highest is \underline{underlined}.}
        \label{tab:main}
    \end{table*}

\section{Experiments}
    Our labeling model has been deployed in the production environment.
    See Appendix \ref{app:deployment} for deployment details.
    
    \subsection{Experimental Setup}
        \noindent \textbf{Content issues}.
            We conduct experiments on the following content issues:
            \textit{Sexually Suggestive Content (SSC)} includes content that contains sexualized framing or sexualized behaviors that are provocative or imply sexual intent.
            \textit{Poor-Quality Content (PQC)} includes videos with static images, excessive on-screen text, lack of audio, or content that provides little to no viewer value or engagement.
            \textit{Clickbait (CB)} includes content that deliberately uses misleading thumbnails, sensationalized titles, or deceptive captions to attract views, but fails to deliver relevant or truthful content as implied.
            \textit{Inappropriate Promotional Content (IPC)} includes promotional material that is misleading, overhyped, or violates advertising standards.

        \vspace{1ex}
        \noindent \textbf{Baseline Methods}.
        We choose LLaVA-OV 7B \citep{li2024llava}, Qwen 2.5-VL 7B\citep{bai2025qwen2}, and GPT-5 \citep{gpt5} as baseline methods for zero-shot short video labeling in L2 manner.

        \vspace{1ex}
        \noindent \textbf{Training and Evaluation Datasets}.
        The training data for the discriminative MLLM labeler comes from human-annotated short videos, with dataset sizes ranging from 1 million to 5 million samples depending on the issue.
        Approximately 10\% of the data consists of positive examples.

        The evaluation datasets contain around 1k to 2k examples, with roughly balanced positive and negative classes.
        For instance, the evaluation set for the SSC issue includes 737 positive and 778 negative samples, while the PQC issue's evaluation set contains 809 positive and 809 negative samples.

        \vspace{1ex}
        \noindent \textbf{Model Details}.
        For L1, the base model is LLaVA \citep{liu2023visual}.
        We fine-tune base models using human-labeled binary video data.
        We apply LoRA-based fine-tuning by adding rank-168 trainable matrices into the transformation matrices of the feed-forward modules at every layer of both the visual encoder and the LLM
        In addition, we add a classification head on top of the final layer.

        For L2, the base model is our pretrained model introduced in Appendix \ref{app:pretrain}.
        We decompose the annotation guideline into sub-items and the prompt is ``\texttt{<Annotation guideline sub-item> <Video info> Does the video violate this rule?}''.
        We aggregate the results for all sub-items.
        If any single sub-item is judged as a violation, the final labeling output is marked as a positive case.
        
        For L3, we implemented the agent using the open-source framework Agno \citep{agno}.
        During each run, we limited the agent to a maximum of 10 tool invocations, with an average of about 5 calls per run.
        The prompt used by the agent is provided in Appendix \ref{app:prompt}.

        \begin{table}[t]
            \centering
            \small
            \setlength{\tabcolsep}{2pt}
            \begin{tabular}{c|ccc}
                \toprule
                \textbf{Type of reduction} & \textbf{SSC} & \textbf{CB} & \textbf{IPC} \\
                \midrule
                \midrule
                Human labor reduction & 27\% & 52\% & 85\% \\
                \midrule
                \multicolumn{1}{c}{Monetary cost reduction} & \multicolumn{3}{c}{} \\
                \makebox[12em][l]{\textit{w/ commercial MLLMs}} & 13--18\% & 25--34\% & 42--56\% \\
                \makebox[12em][l]{\textit{w/ locally deployed MLLMs}} & 25\% & 48\% & 79\% \\
                \bottomrule
            \end{tabular}
            \caption{Human labor reduction and monetary cost reduction after introducing MLLM labeler.}
            \label{tab:cost}
        \end{table}

            \begin{figure*}[t]
                \begin{subfigure}{0.3\textwidth}
                    \centering
                    \includegraphics[width=\linewidth]{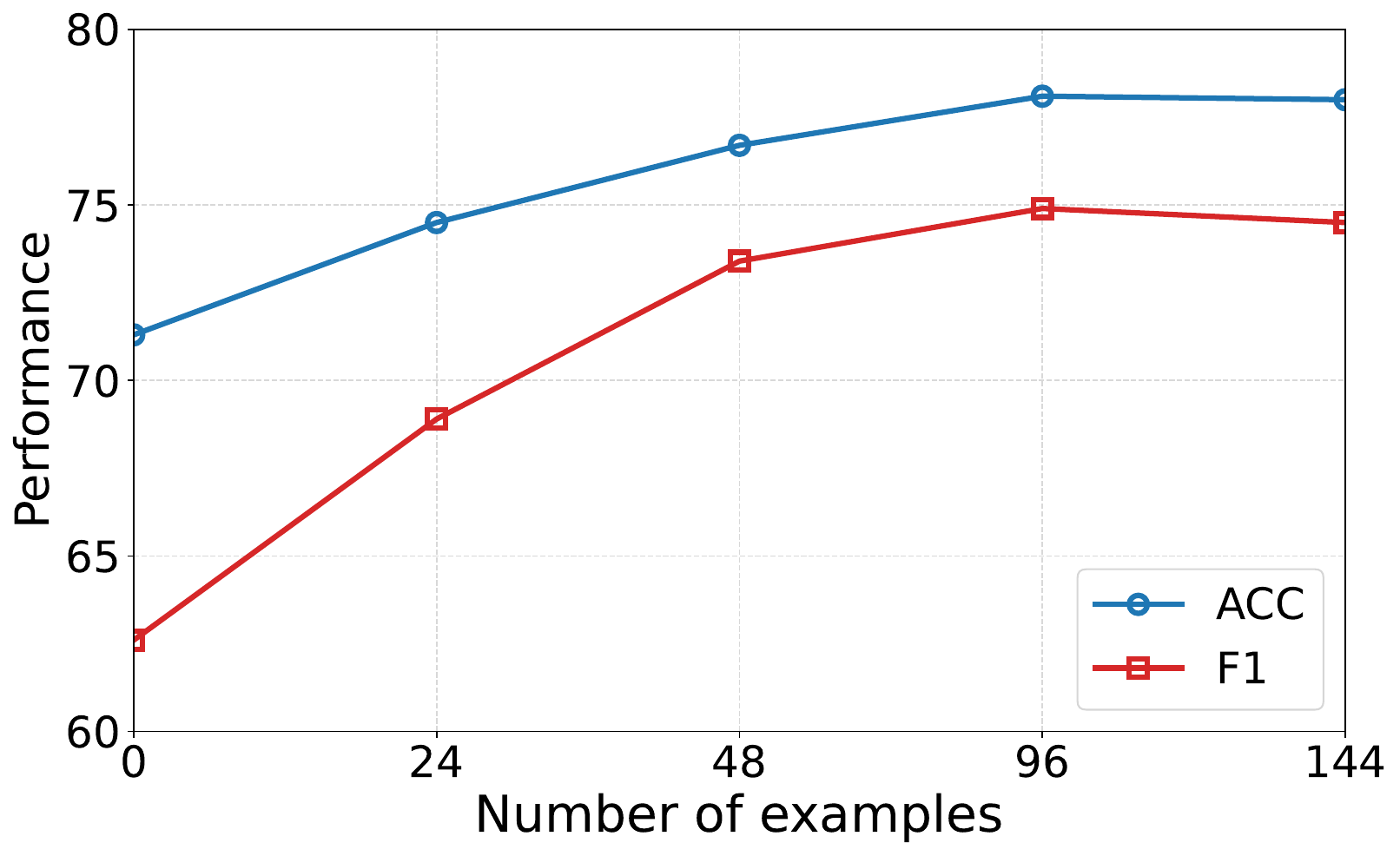}
                    \caption{SSC}
                    \label{fig:example_nums_ssc}
                \end{subfigure}
                \hfill
                \begin{subfigure}{0.3\textwidth}
                    \centering
                    \includegraphics[width=\linewidth]{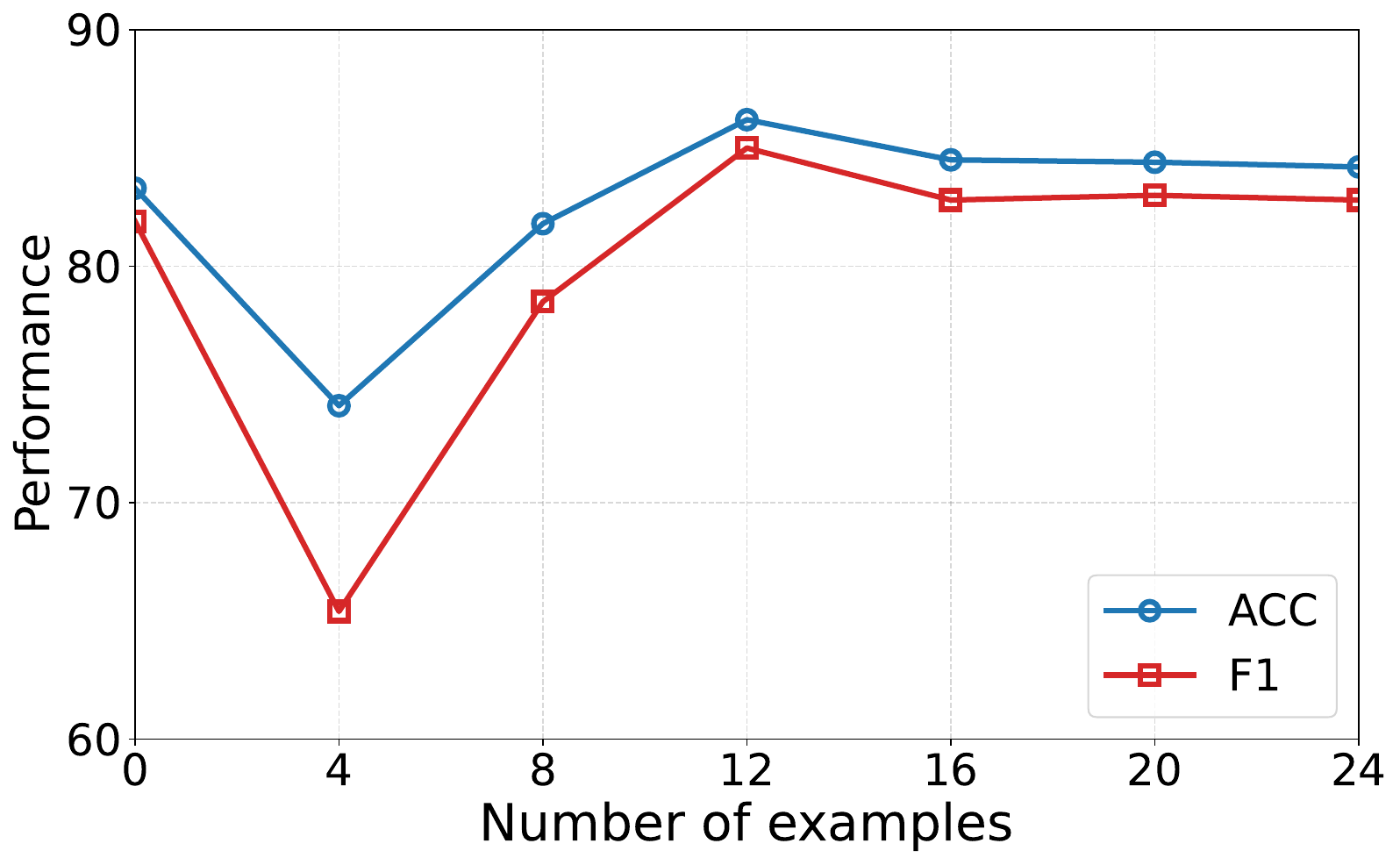}
                    \caption{PQC}
                    \label{fig:example_nums_pqc}
                \end{subfigure}
                \hfill
                \begin{subfigure}{0.3\textwidth}
                    \centering
                    \includegraphics[width=\linewidth]{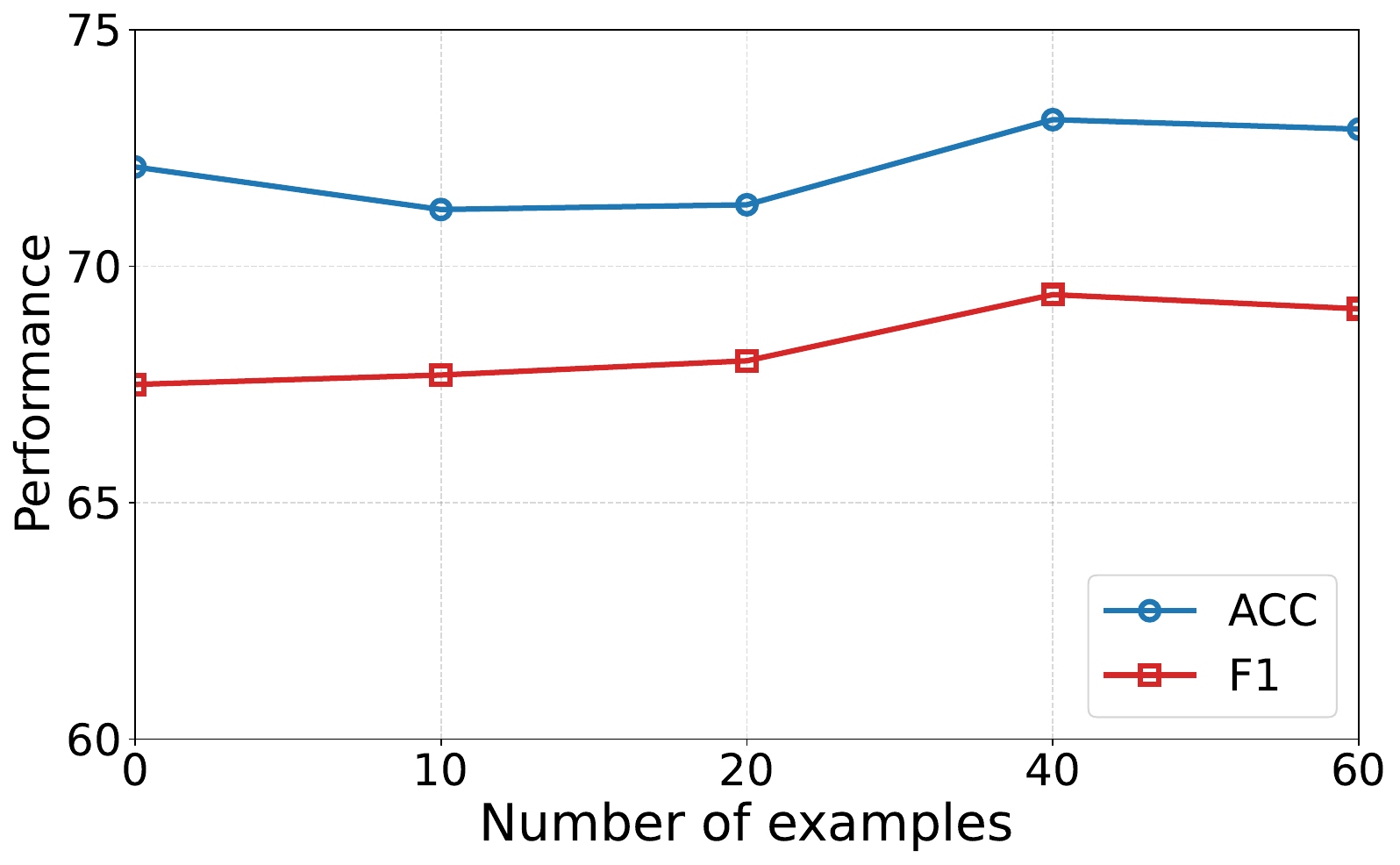}
                    \caption{CB}
                    \label{fig:example_nums_cb}
                \end{subfigure}
                \caption{Impact of the number of examples on agent's performance.}
                \label{fig:example_nums}
            \end{figure*}

    \subsection{Main Results}
    \label{sec:main_results}
        \noindent \textbf{Comparison between labeling models}.
            The results of all labeling models are presented in Table \ref{tab:main}.
            The number of precedents the agent can refer to for SSC, PQC, and CB is 96, 12, and 40, respectively.
            Overall, the experimental results demonstrate that as the level of automation in labeling increases (from L1 to L3), model performance improves accordingly.
            The row of ``Combining L1 and L3'' refers to a hybrid approach: using the discriminative labeler from the L1 stage for initial predictions, and then re-labeling uncertain cases (those with non-extreme prediction probabilities) using the L3-stage agent. This result suggests that combining discriminative and generative MLLM labelers can further boost performance.
            We also calculate the \textit{confidence intervals} for our best method (combining L1 and L3). The 95\% confidence intervals of their F1 scores are as follows. SSC: 81.3 $\pm$ 2.2, PQC: 95.8 $\pm$ 1.5, CB: 74.6 $\pm$ 1.7, which show significantly improvement of our best method.

        \vspace{1ex}
        \noindent \textbf{Human annotation cost reduction}.
            The details of how human and MLLM collaborate during deployment is in Appendix \ref{app:experiment_collaboration}.
            We report the human labor reduction achieved after deploying the MLLM labeler, as shown in Table \ref{tab:cost}.
            For SSC, CB, and IPC issues, human labor reduction is 27\%, 52\%, and 85\%, respectively.
            Assuming the monetary cost of human annotation for each video is 1 unit, then the cost per video using commercial MLLMs is 0.34--0.51 units, while using locally deployed MLLMs reduces the cost to just 0.068 units per video.
            Based on these numbers, we calculated the monetary cost reduction achieved by introducing MLLM labelers, as summarized in Table \ref{tab:cost}.
            The hybrid approach combining human annotators and locally deployed MLLMs can reduce annotation costs by 25\% to 79\%.

            \begin{figure}[t]
                \begin{subfigure}{0.23\textwidth}
                    \centering
                    \includegraphics[width=\linewidth]{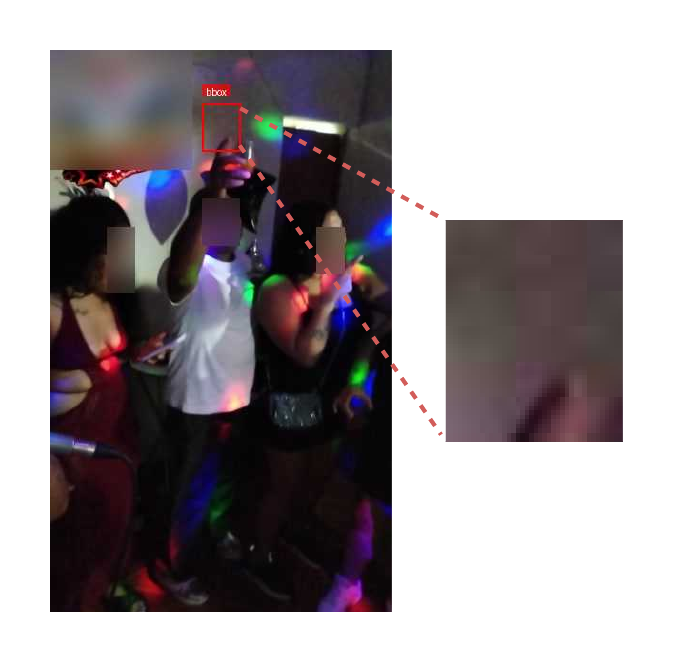}
                    \caption{Bounding box generation}
                    \label{fig:zoomin_1}
                \end{subfigure}
                \hfill
                \begin{subfigure}{0.23\textwidth}
                    \centering
                    \includegraphics[width=\linewidth]{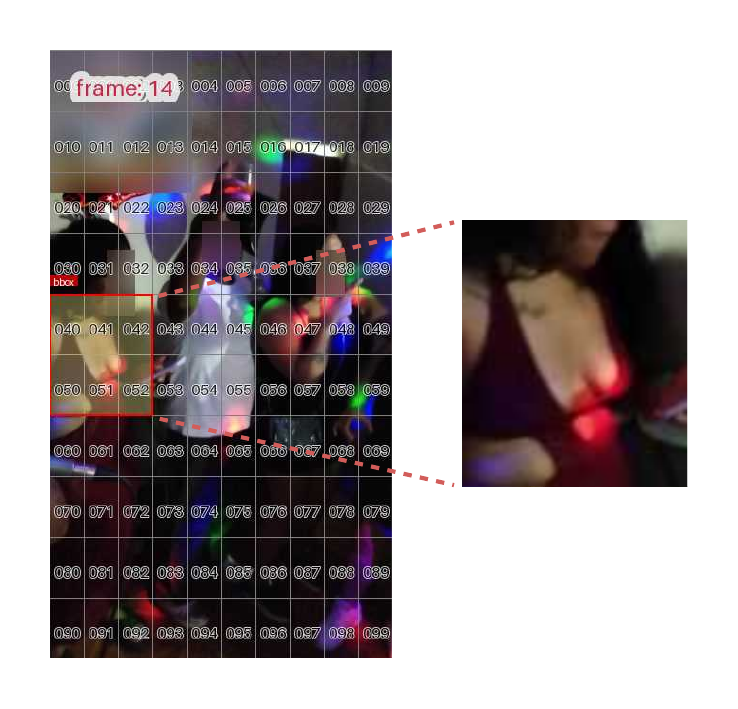}
                    \caption{Visual grid selection}
                    \label{fig:zoomin_2}
                \end{subfigure}
                \caption{The two implementations of the zoom-in tool.}
                \label{fig:zoomin}
            \end{figure}

    \subsection{Ablation Study}
    \label{sec:ablation}   
        \vspace{1ex}
        \noindent \textbf{Impact of the number of examples}.
            We vary the number of examples and see how the agent's performance changes.
            The result for SSC, PQC, and CB are shown in Figure \ref{fig:example_nums}.
            The case with 0 examples corresponds to the zeroshot setting.
            As shown, the agent performs poorly when given no example or only a small number of examples.
            When the number of examples is too large, performance saturates and even shows a declining trend.
            This confirms the observations discussed in Section \ref{sec:l3}.

            The difference in the optimal number of precedents across issues is due to the significant difference in the complexity of their annotation guidelines.
            The number of sub-issues for SSC, PQC, and CB is 17, 4, and 12, respectively.
            Since each sub-issue requires precedents, the total number of precedents for SSC is therefore the highest, while for PQC is the smallest.

            \begin{table}[t]
                \centering
                \setlength{\tabcolsep}{6pt}
                \small
                \begin{tabular}{c|cccc}
                    \toprule
                    \textbf{Zoom-in tool} & \textbf{Acc} & \textbf{P} & \textbf{R} & \textbf{F1} \\
                    \midrule
                    \midrule
                    No zoom-in & 78.3 & 86.2 & \textbf{68.2} & 76.1 \\
                    \midrule
                    Bounding box generation & 77.8 & 87.7 & 65.4 & 74.9 \\
                    \midrule
                    Visual grid selection & \textbf{79.0} & \textbf{88.2} & 67.5 & \textbf{76.5} \\
                    \bottomrule
                \end{tabular}
                \caption{Comparison of zoom-in tools on the SSC issue. The highest value in each column is shown in \textbf{bold}.}
                \label{tab:zoomin}
            \end{table}
        
        \vspace{1ex}
        \noindent \textbf{Different implementations of the zoom-in tool}.
            Figure \ref{fig:zoomin} illustrates the two zoom-in strategies discussed in Section \ref{sec:l3}.
            The first approach often suffers from drift, making it difficult to accurately select the correct region.
            In contrast, the second strategy can reliably localize the exposed chest area.
            Their performance is shown in Table \ref{tab:zoomin}.
            Bounding box generation performs worse than visual grid selection, and even underperforms compared to the no zoom-in baseline.

            \begin{figure}[t]
                \begin{subfigure}{0.5\textwidth}
                    \includegraphics[width=0.9\linewidth]{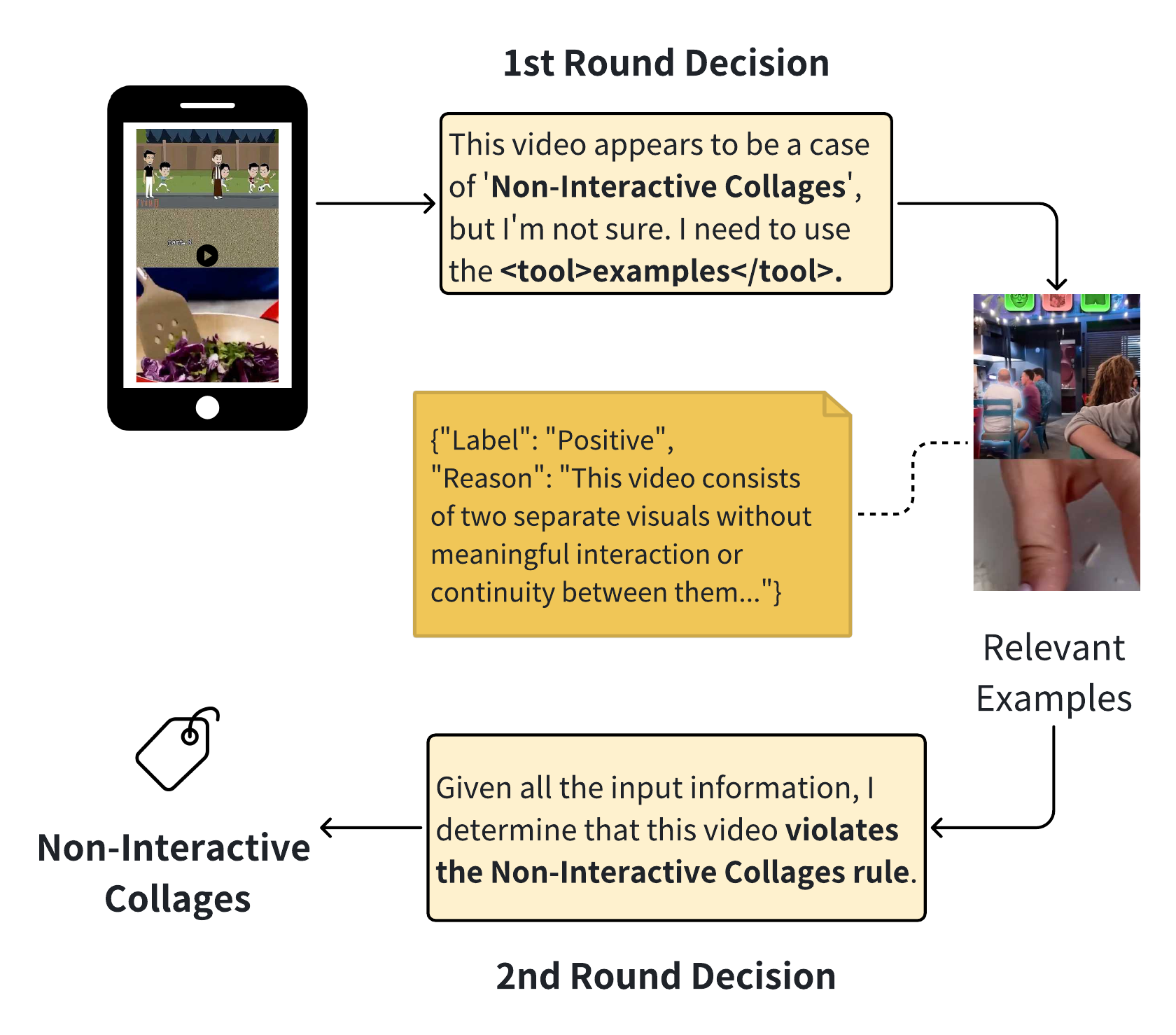}
                    \caption{PQC}
                    \label{fig:case_study_1}
                \end{subfigure}
                \hfill
                \begin{subfigure}{0.5\textwidth}
                    \vspace{2em}
                    \includegraphics[width=0.9\linewidth]{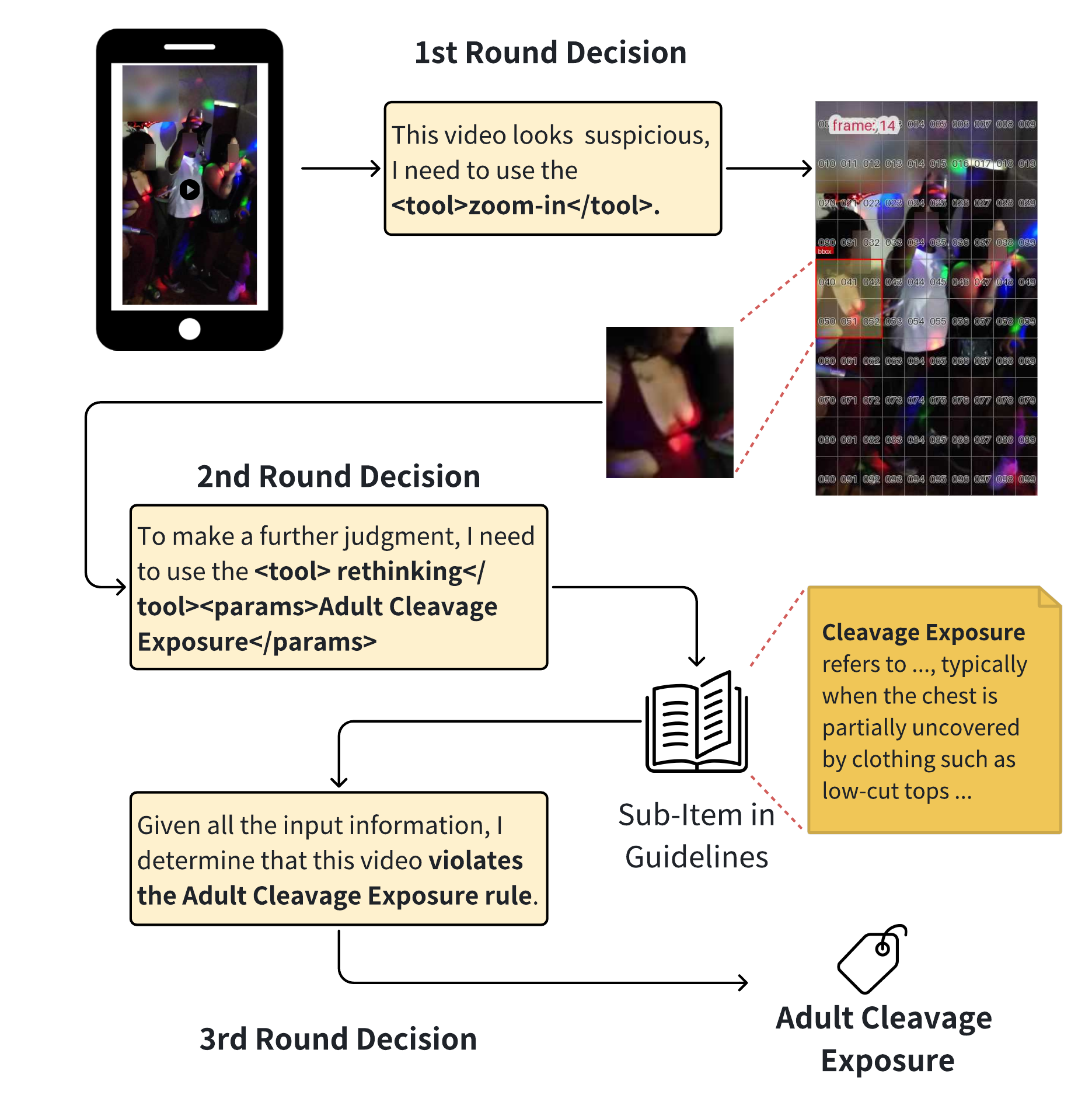}
                    \caption{SSC}
                    \label{fig:case_study_2}
                \end{subfigure}
                \caption{Success case analysis.}
                \label{fig:case_study}
            \end{figure}

    \subsection{Case Analysis}
    \label{sec:case}
        \noindent \textbf{Success Case Analysis}.
        We conduct two case studies to demonstrate the importance of the three tools integrated into the agent.
        The corresponding images are shown in Figure \ref{fig:case_study}.
        Figure \ref{fig:case_study_1} focuses on PQC issue, where the input video is a combination of two unrelated clips. The agent is initially uncertain about its judgment, so it invokes the example tool.
        After referencing similar examples, it confidently determines the correct answer and produces the output.
        Figure \ref{fig:case_study_2} addresses an SSC issue involving a blurry and chaotic nightclub scene, in which there is a brief instance of breast exposure.
        Since the exposed area is small, the agent first use the zoom-in tool to enlarge the relevant detail, filtering out distracting noise from the rest of the scene.
        It then calls the rethinking tool to review the precise definition of the relevant sub-item.
        Based on this process, the agent arrives at a final conclusion.

        \vspace{1ex}
        \noindent \textbf{Failure Case Analysis}.
        In summary, the failure cases mainly fall into two categories:
        (1) Recognition of continuous actions, such as piston movements and hip-thrust actions in SSC issue. These fast-paced continuous motions are difficult to recognize under the current frame-sampling strategy. We plan to address this by adopting a more intelligent frame sampling approach.
        (2) Audio-related issues, such as silence (i.e., videos with no sound) in PQC issue, and moaning sounds in SSC issue. We plan to incorporate audio understanding tools to address these problems.

\section{Conclusion and Future Work}
    This paper presents our progress in automatic short video labeling.
    Inspired by the levels of autonomous driving, we propose a four-stage technical roadmap: manual labeling (L0) → discriminative MLLM labeling (L1) → generative MLLM labeling (L2) → agentic labeling (L3).
    Experimental results demonstrate that methods along this roadmap progressively reduce reliance on manual labor, while achieving improvements in annotation performance, cost reduction, and generalization to complex or unseen cases.

    Looking ahead, a fully autonomous (L4) labeler requires two advances.
    First, stronger reasoning in the base model, which we plan to pursue through better-designed domain-adaptive pretraining tasks and higher-quality pretraining data.
    Second, a richer agent toolset: beyond the intelligent frame sampling and audio
    understanding noted in the Limitations, we plan to add external information lookup
    (e.g., querying the Internet to check whether a video is a clip from a recent film) and retrieval over a large precedent database, so that the agent selects its own references instead of relying on a fixed example set per sub-item.

\section*{Limitations}
    We discuss three main limitations of our work, along with their practical impact and potential mitigation.

    \vspace{1ex}
    \noindent \textbf{Perception bottlenecks in frames and audio}.
        Our vision-centric pipeline represents each video as uniformly sampled frames, leading to two failure patterns (Section \ref{sec:case}): fast continuous actions (e.g., piston movements in SSC) whose cues fall between sampled frames, and audio-dependent violations (e.g., silence in PQC, moaning sounds in SSC) that are invisible without audio modeling.
        These cases are relatively rare and have limited impact on overall performance. Moreover, both are naturally addressable within our agentic design by plugging in intelligent frame-sampling and audio-understanding tools in future work.

    \vspace{1ex}
    \noindent \textbf{Computational overhead of the agent}.
        The L3 agent trades efficiency for accuracy, issuing about five times more queries per video and raising the daily cost from 11.1 to 83.5 H100 GPU-hours, with higher latency than L1/L2 (Appendix \ref{app:agentic_discussion}).
        This overhead may constrain applicability under tight latency or budget requirements. In our offline labeling setting, however, it stays well within deployment capacity, and the hybrid scheme further reduces the average cost by invoking the agent only on uncertain cases.

    \vspace{1ex}
    \noindent \textbf{Reliance on human-labeled and expert resources}.
        The framework still depends on human input: the L1 labeler requires large amounts of supervised data with performance improving as more is added (Appendix \ref{app:experiment_ablation}), and later stages rely on expert-curated precedents and expanded guidelines.
        It thus remains a human-MLLM collaboration rather than a fully autonomous labeler. Nevertheless, this dependence drops sharply from L1 to L3, and the resources are largely one-time: once prepared, they are reused long-term, so the marginal annotation cost keeps decreasing as more videos are processed.

\section*{Ethics Statement}
    Labeling errors in content governance can reduce a creator's reach or lead to removal.
    We summarize how the deployed system is scoped, overseen, and audited.

    \vspace{1ex}
    \noindent \textbf{Scope of model-generated labels}.
        Model labels are used to build training data for the downstream governance models (Appendix \ref{app:experiment_compare_governance}) and to measure issue prevalence.
        They are \textit{not} a standalone trigger for user-facing enforcement: any visibility reduction or removal is issued by the separately deployed governance pipeline, which retains its own review and appeal stages.

    \vspace{1ex}
    \noindent \textbf{Human oversight}.
        The deployment is human-in-the-loop by construction (Figure \ref{fig:mllm_labeling}): only predictions above $\Delta$ or below $\delta$ are auto-accepted (Appendix \ref{app:experiment_collaboration}), all cases in between are escalated to human annotators, and the MLLM never forms a majority on its own in dual review.
        Auto-accepted labels are QA-sampled at the same rate as human labels, and labels corrected on appeal are fed back as supervision.

    \vspace{1ex}
    \noindent \textbf{Fairness}.
        Automated moderation can encode cultural bias, e.g., by reading region-specific dress or vernacular as a violation.
        Disaggregating the L3 agent's errors on the SSC evaluation set by the primary language of the video's text signals, F1 ranges from 72.4 to 79.1 across the five largest language groups (76.5 overall), and errors in the weakest group skew toward false positives (FPR 13.2\% vs.\ 8.6\%), the more harmful direction for creators.
        We mitigate this by curating the precedent pool to span the regions the platform serves; a broader audit across intersectional slices remains future work and a prerequisite for raising the auto-accepted share.

    \vspace{1ex}
    \noindent \textbf{Privacy}.
        Videos may contain faces, voices, and private settings.
        The deployed labelers run on internal infrastructure and send no user video to third-party APIs; the GPT-5 baseline (Table \ref{tab:main}) was run only on our held-out evaluation set, and the commercial-MLLM figures (Table \ref{tab:cost}) are cost projections.
        Access is restricted, logged, and governed by the platform's retention and privacy policies, and precedents are added only through expert curation.

    \vspace{1ex}
    \noindent \textbf{Annotator well-being and transparency}.
        Absorbing the high-confidence workload reduces annotators' exposure to distressing material. 
        The agent's rationales are currently used internally for QA; surfacing them to affected creators is future work.

\bibliography{reference}

\clearpage
\appendix

\section{More Discussion on Manual Labeling}
    \label{app:l0}
\begin{figure}[t]
        \begin{subfigure}{0.5\textwidth}
            \centering
            \includegraphics[width=0.6\linewidth]{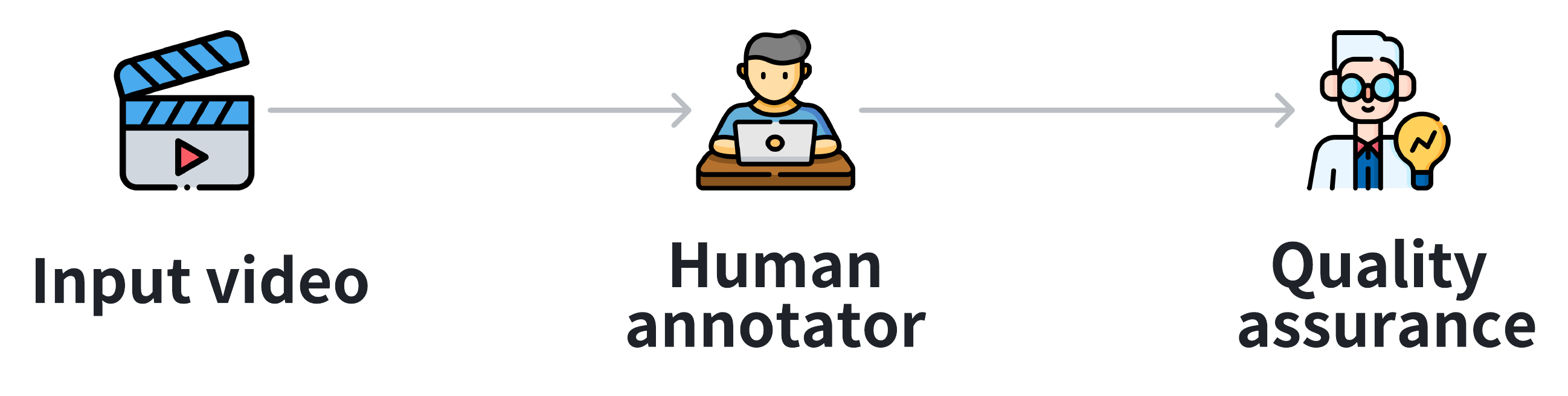}
            \caption{Single-review annotation}
            \label{fig:manual_labeling_1}
            \vspace{1em}
        \end{subfigure}
        \vfill
        \begin{subfigure}{0.5\textwidth}
            \centering
            \includegraphics[width=\linewidth]{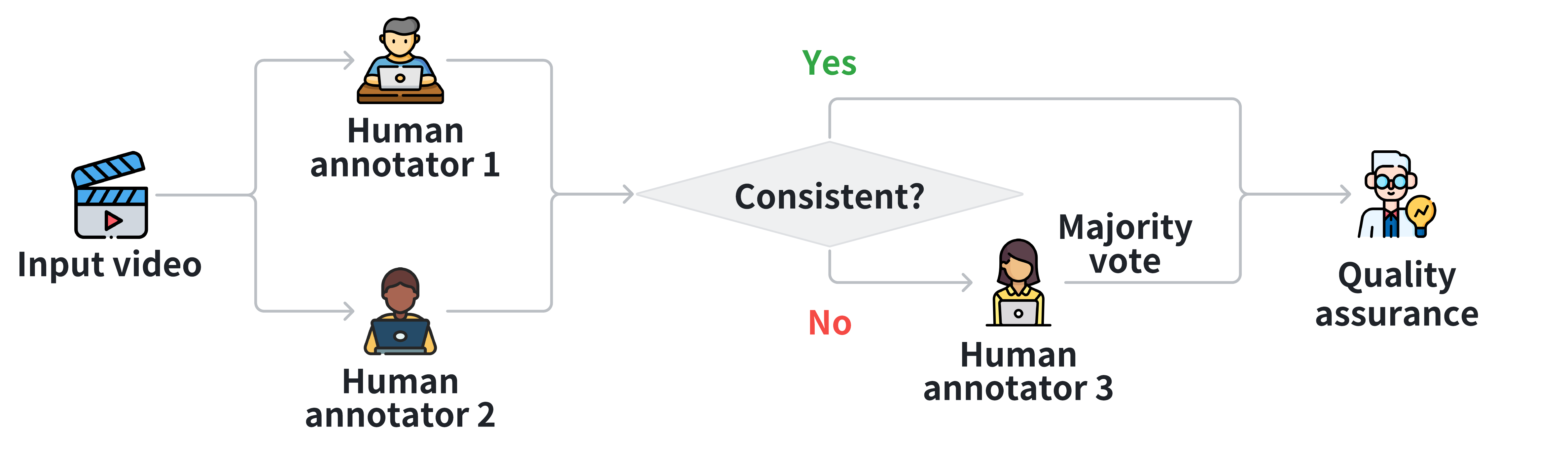}
            \caption{Dual-review annotation}
            \label{fig:manual_labeling_2}
        \end{subfigure}
        \caption{The work flows of manual labeling.}
        \label{fig:manual_labeling}
    \end{figure}

    As illustrated in Figure \ref{fig:manual_labeling}, the manual annotation workflow for short videos on our platform can be categorized into two types depending on the complexity of the labeling task:
    
    \vspace{1ex}
    \noindent \textbf{Single-review annotation}.
    For those issues with clearly defined guidelines and straightforward judgments (e.g., clickbait), each video is reviewed and labeled by a single annotator.
    The results are subject to random sampling and re-evaluation by quality assurance (QA) personnel to ensure accuracy.
    
    \vspace{1ex}
    \noindent \textbf{Dual-review annotation}.
    For more complex or subjective issues (e.g., sexually suggestive content), each video is independently labeled by two annotators.
    If both annotators produce the same label, that result is accepted.
    If their labels differ, a third annotator is brought in, and the final label is determined by majority vote.

    \vspace{1em}
    The limitations of manual labeling are:
    \begin{itemize}
        \item
            \textit{Low accuracy and consistency}.
            The quality of manual annotations is often limited by the annotators' education level and experience.
        \item
            \textit{Slow and inefficient}.
            Manual labeling is time-consuming.
            Annotators must watch each short video (ranging from a few seconds to several minutes) and carefully apply complex annotation guidelines.
            Furthermore, each video may require multiple reviewers, multiplying the total annotation time.
        \item
            \textit{High operational cost}.
            The financial burden of manual labeling is substantial, with monthly annotation costs reaching tens of millions of dollars.
        \item
            \textit{Long iteration cycles}.
            In practice, annotation guidelines are frequently updated, often on a monthly basis.
            Each update requires re-training of annotators before any new data can be produced, significantly delaying the labeling cycle.
    \end{itemize}

\section{More Discussion on Discriminative MLLM labeling}
    \label{app:l1}
    It's worth noting that this discriminative MLLM labeler is technically similar to the models used for online content governance.
    However, there are two key differences:
    (1) \textit{Purpose}.
    The labeling model is used to generate classification labels for videos, which are then used to train online governance models.
    (2) \textit{Model size}.
    Online models must meet strict queries-per-second (QPS) requirements and are therefore typically small, just a few hundred megabytes.
    In contrast, the labeling model is deployed offline and can afford to be much larger in size to maximize performance.
    The comparison between discriminative MLLM labeler and governance models in terms of experimental result is in Appendix \ref{app:experiment_compare_governance}.

    One concern is that discriminative MLLM labelers still require supervised training, which in turn relies on human-annotated data.
    This raises the question: do they truly save on manual labeling costs?
    Our answer is YES.
    While initial training does require human-labeled data, once the model is trained, it can take over a substantial portion (or even the majority) of future labeling work.
    This is a classic case of ``\textit{train once, benefit long-term}''.

\section{Details on Domain-Adaptive Continued Pretraining}
    \label{app:pretrain}
        While native MLLMs generally exhibit strong generalization capabilities, applying them to the short video annotation scenario on our platform still presents two key challenges:
        (1) \textit{Domain-specific data distribution}.
        Compared to the data used during the MLLM's pretraining, short videos have unique characteristics in both format and semantics.
        This includes distinctive frame layouts, editing styles, and the use of internet-specific language and slang.
        (2) \textit{Complex issue definitions and annotation guidelines}.
        Native MLLMs are not exposed to these domain-specific concepts during pretraining.
        However, to perform labeling tasks effectively, the model must not only understand but also strictly adhere to these complex and often evolving annotation guidelines.

        To address these challenges, we conduct domain-adaptive continued pretraining on MLLMs.
        We specifically designed three pretraining tasks:
        \begin{itemize}
            \item
            \textbf{Captioning}.
            This task aims to train MLLMs to describe the content of a video, especially those related to specific issues.
            For example, a prompt might be ``\texttt{Describe the video regarding sexually suggestive content.}''
            This can help the model learn to focus on fine-grained details in the video, thereby enhancing its perceptual capabilities.
            
            \item
            \textbf{VQA} (Visual Question Answering).
            To ensure the model thoroughly understands annotation guidelines, we construct a set of questions and corresponding answers based on these guidelines.
            For instance, for the issue of sexually suggestive content, questions include ``\texttt{Is private body part exposed?}'' and ``\texttt{Is there sexual teasing/invitation?}''
            This allows the model to internalize and apply the annotation criteria more effectively.
            
            \item
             \textbf{CoT} (Chain of Thought).
             Even after perceiving video details and understanding annotation guidelines, the model still needs to follow a systematic reasoning process to reach a final decision.
             For example, in ``non-original content'' issue, the reasoning might involve steps like: ``\texttt{First, check if the video contains watermarks... Then, determine whether it includes copyrighted materials such as TV shows... Next, count the number of clips... Based on these factors, the video is plagiarized/original.}''
             To support this, we designed a CoT task that teaches the model to reason step by step following predefined logical flows.
        \end{itemize}

        Based on the three pretraining tasks, we propose three pretraining strategies:
        \begin{itemize}
            \item
            \textbf{Caption-only}.
            In this strategy, the language model is frozen while the vision encoder and projector are trained using only the caption dataset.
            The goal is to enhance the model’s ability to perceive fine-grained visual details.
    
            \item
            \textbf{Mixing-all-tasks}.
            All three types of pretraining datasets are mixed and randomly shuffled.
            We train all components of the MLLM jointly on this mixed dataset.
    
            \item
            \textbf{Two-stage}.
            This is a two-stage approach that combines the previous two strategies.
            We first apply Caption-only strategy to train the vision encoder and projector, and select the best checkpoint across epochs.
            Then, we switch to the Mixing-all-tasks strategy to fine-tune the entire model.
        \end{itemize}

    The experimental results demonstrate that, Two-stage consistently outperforms Caption-only and Mixing-all-tasks in most cases.
    This suggests that first training the model to perceive fine-grained video details, followed by learning the annotation guidelines and reasoning process, is the most effective practice.

    We conduct the domain-adaptive pretraining on LLaVA \citep{liu2023visual} using a total of 920k instruction samples from three content issues: sexually suggestive content, unoriginal content, and clickbait.
    The model obtained through this continued pretraining will serve as our MLLM labeler.

\section{More Analysis and Discussion on Agentic Labeling}
\label{app:agentic_discussion}
    \subsection{Benefits and Challenges}
        Deploying automatic labeling for videos brings several advantages.
        It significantly improves scalability and efficiency, allowing large volumes of videos to be reviewed automatically.
        Labeling models can operate continuously, reduce labeling time, and ensure faster turnaround for content governance.
        In addition, automated systems can maintain more consistent standards and be easily fine-tuned as policies evolve.
        
        However, transitioning from human to auto labeling introduces some challenges.
        Models may struggle with ambiguous or context-dependent content, leading to false positives or missed violations.
        Moreover, deployment involves technical challenges such as integration with existing pipelines and inference cost at scale.

    \subsection{Computational Cost}
        We compare the computational resource costs of discriminative MLLM labeler (L1) and agent (L3).
        On average, we annotate about 200k videos per day. The L1 model achieves a QPS of 5 on H100 GPUs, which means it requires approximately 11.1 H100 GPU-hours per day.
        The L3 model generates around five times more queries per video, and each query contains about 1.5 times more tokens.
        As a result, the agent requires about 83.5 H100 GPU-hours per day.
        Although the agent demands more computational resources, the additional 83.5 GPU-hours per day do not impose any noticeable strain on our deployment.

    \subsection{Latency}
        The labeling latency for each short video is as follows:
        \begin{itemize}
            \item L0 (human): 35.1 – 80.4s;
            \item L1 (discriminative MLLM labeler): 1.5 - 1.8s;
            \item L2 (generative MLLM labeler): 5.5 - 12.2s;
            \item L3 (agent): 10.5 – 30.1s.
        \end{itemize}
        The agent's latency is higher than that of L1/L2, but it is still significantly better than human labeling.

    \subsection{Generalization}
        Our video labeling models often demonstrate strong generalization capabilities across various video annotation tasks.
        Because they are trained on large-scale, diverse datasets that cover multiple scenes, actions, and visual contexts, these models can effectively transfer their learned representations to related tasks such as activity recognition, content tagging, and highlight detection.
        Their ability to capture temporal patterns and semantic relationships enables them to adapt with minimal fine-tuning, reducing the need for task-specific data.
        This cross-task flexibility not only improves efficiency but also accelerates the development of new video understanding applications.

\section{System Deployment}
    \label{app:deployment}

    \begin{figure}[t]
        \begin{subfigure}{0.47\textwidth}
            \centering
            \includegraphics[width=\linewidth]{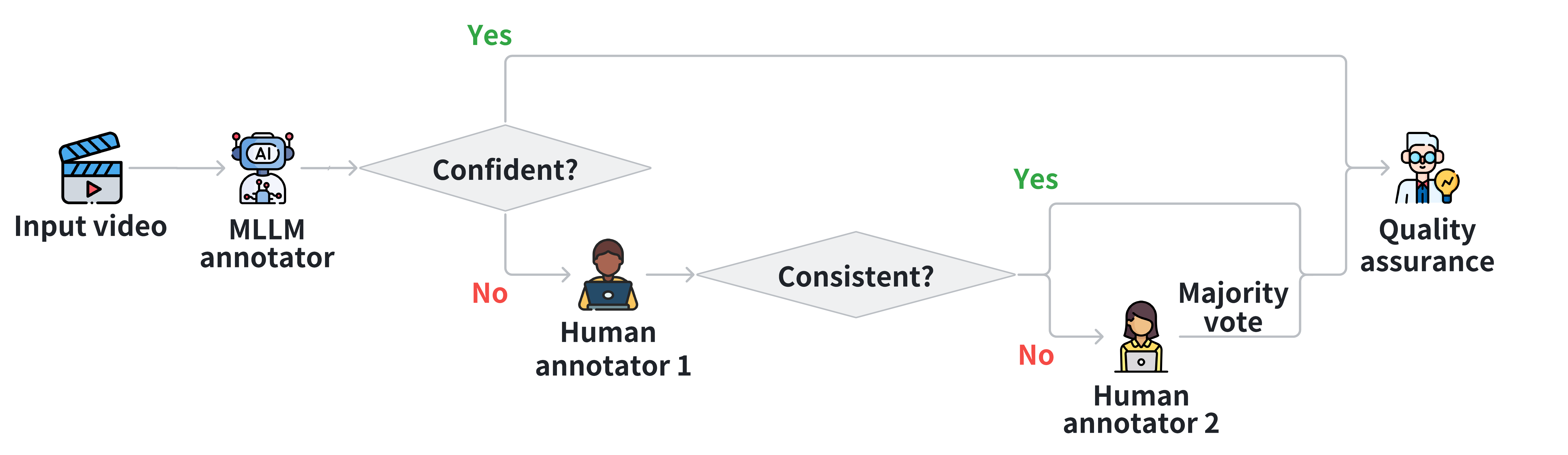}
            \caption{MLLM assisted single-review annotation}
            \label{fig:mllm_labeling_1}
            \vspace{1em}
        \end{subfigure}
        \vfill
        \begin{subfigure}{0.41\textwidth}
            \centering
            \includegraphics[width=\linewidth]{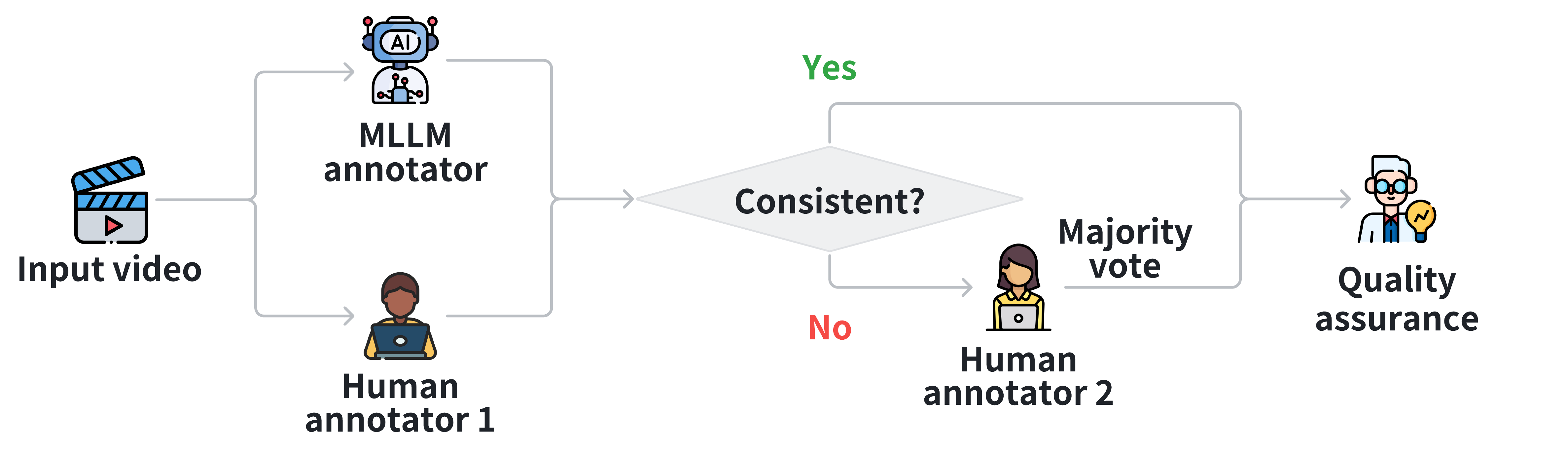}
            \caption{MLLM assisted dual-review annotation}
            \label{fig:mllm_labeling_2}
        \end{subfigure}
        \caption{The work flows of MLLM-assisted labeling.}
        \label{fig:mllm_labeling}
    \end{figure}

    Our labeling model has been deployed in the production environment, as illustrated in Figure \ref{fig:mllm_labeling}.

    \vspace{1ex}
    \noindent \textbf{MLLM-assisted single-review annotation}.
    In Figure \ref{fig:mllm_labeling_1}, we replace the human annotator in Figure \ref{fig:manual_labeling_1} with an MLLM annotator.
    However, relying entirely on the MLLM is not yet practical.
    Therefore, we adopt a confidence-aware strategy:
    If the model is highly confident in its labeling result, we accept it directly.
    Otherwise, a human annotator is brought in:
    If the MLLM and the human annotator agree, the label is accepted;
    If they disagree, a second human annotator is introduced, and the final result is determined by majority vote.

    \vspace{1ex}
    \noindent \textbf{MLLM-assisted double-review annotation}.
    In Figure \ref{fig:mllm_labeling_2}, we replace one of the human annotators in Figure \ref{fig:manual_labeling_2} with an MLLM annotator, while keeping the rest of the process unchanged.

    \vspace{1ex}
    An interesting observation is that both of the above cases involve one MLLM annotator and up to two human annotators.
    Why not use two or more MLLM annotators instead to further reduce labor costs?
    The reason lies in the high consistency of MLLM annotations: adding a second or third MLLM often results in the same label as the first annotator.
    This undermines the purpose of majority voting and offers little benefit in improving label robustness.

\section{More Experimental Results}

        \begin{figure}[t]
            \centering
            \includegraphics[width=0.35\textwidth]{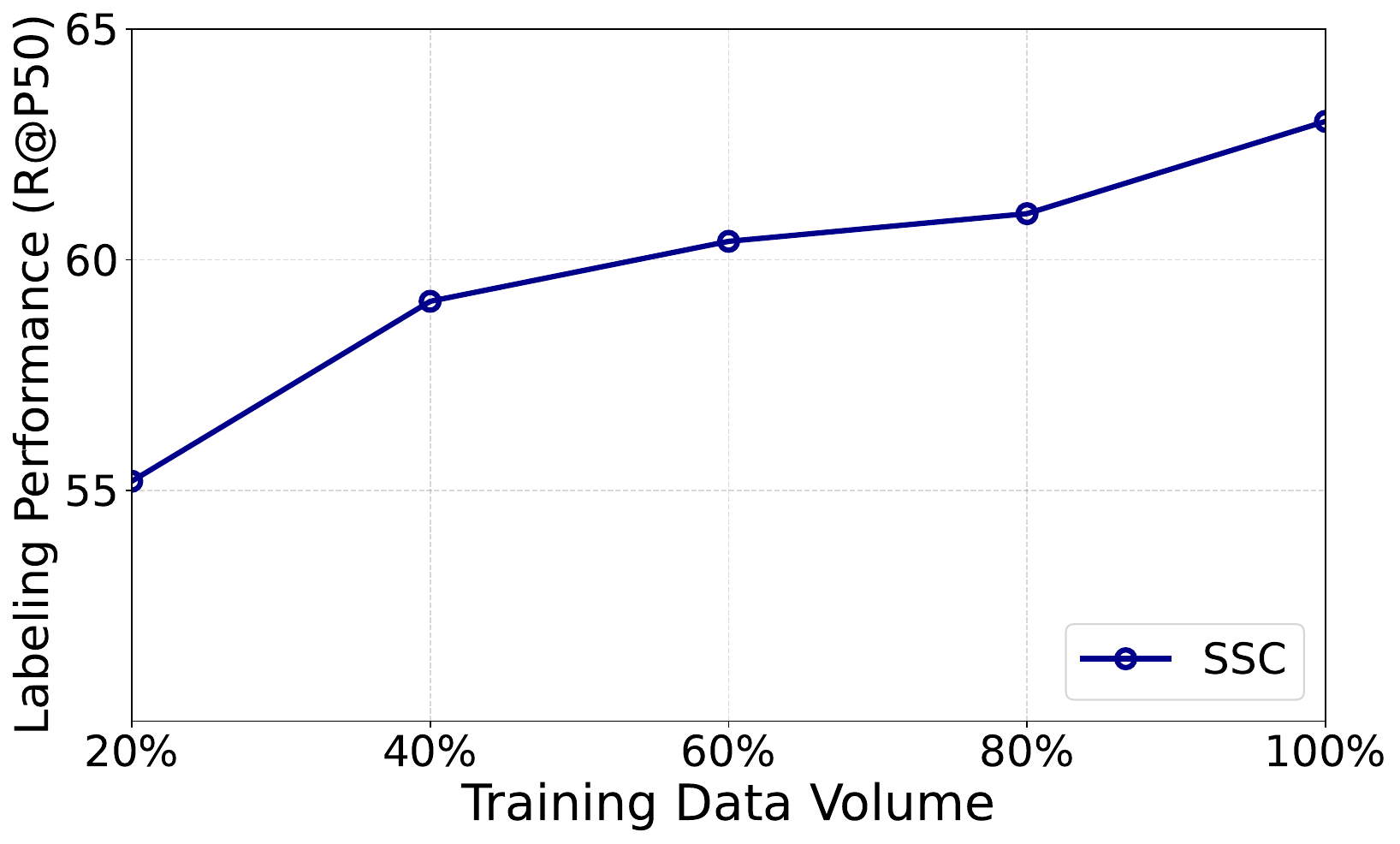}
            \caption{Discriminative MLLM's labeling performance on the SSC issue with varying amounts of training data.}
            \label{fig:sft_training_data}
        \end{figure}

    \subsection{Impact of Training Data Volume on the Labeling Performance of Discriminative MLLMs}
    \label{app:experiment_ablation}
        The discriminative MLLM labeler requires supervised data for training.
        To investigate how the amount of training data affects its labeling performance, we gradually reduce the training data for the SSC issue from 100\% to 20\%, train the model at each level, and evaluate its performance.
        The results are presented in Figure \ref{fig:sft_training_data}, which shows that the labeling performance steadily improves with more training data: R@P50 increases from 55.2 to 63.0, and continues to show an upward trend.
        On the positive side, this indicates that the performance of the discriminative MLLM labeler still has room for improvement.
        However, it also highlights a significant drawback: it relies heavily on large amounts of manually labeled data to achieve better performance.
        From this perspective, the generative MLLM labeler has a clear advantage in terms of cost efficiency.

        \begin{figure}[t]
            \centering
            \includegraphics[width=0.45\textwidth]{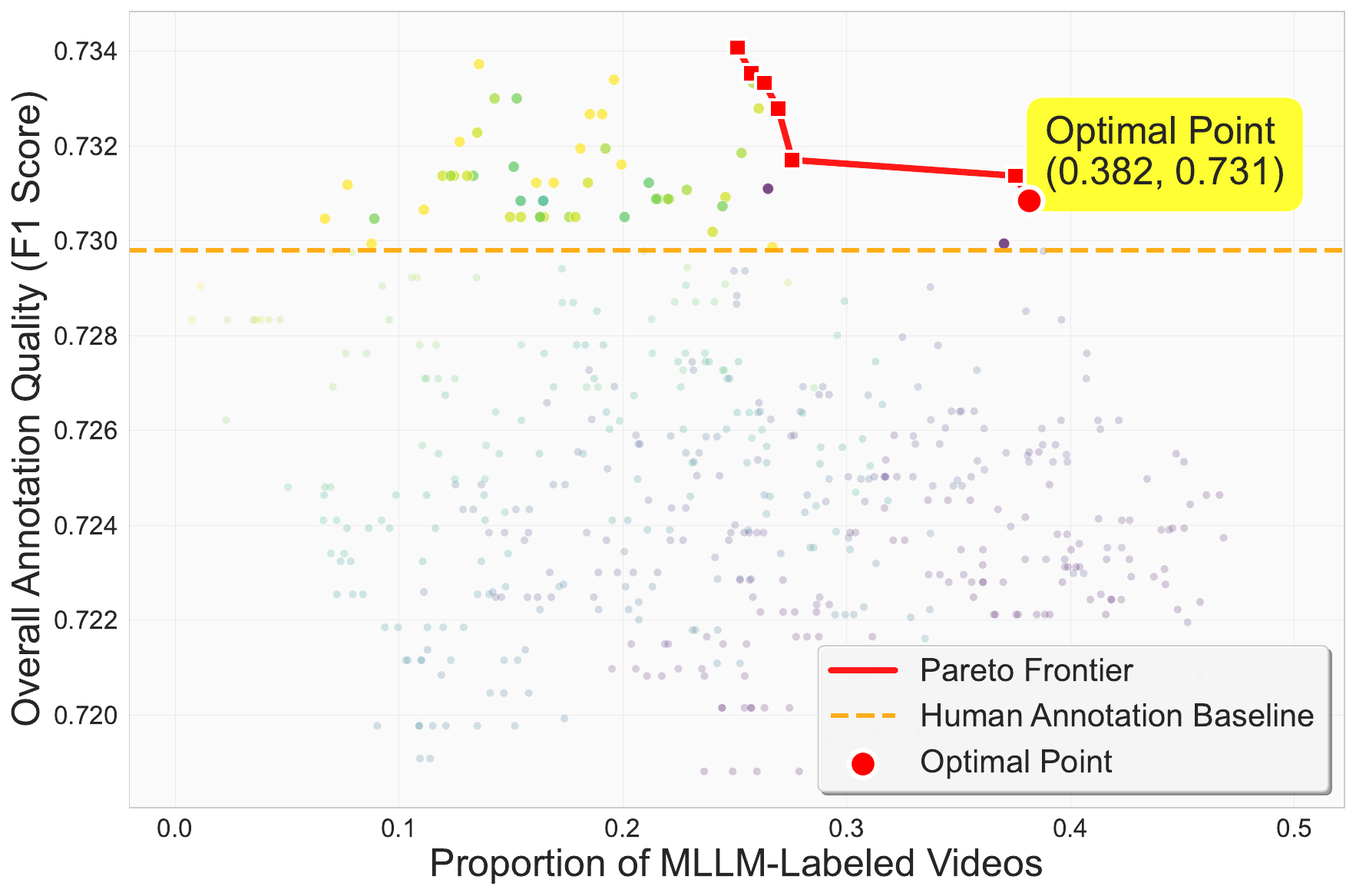}
            \caption{The correlation between the proportion of MLLM-labeled videos and the overall annotation quality for SSC (See Section \ref{sec:main_results} for details). The optimal point indicates that up to 38.2\% of human labor can be saved theoretically.}
            \label{fig:cost}
        \end{figure}
        
    \subsection{Human-MLLM Collaboration}
    \label{app:experiment_collaboration}
        Current deployment uses a human-MLLM collaborative approach.
        As shown in Figure \ref{fig:mllm_labeling_1}, only model predictions with high confidence are accepted, i.e., when the predicted probability of being a positive example is above an upper threshold $\Delta$ or below a lower threshold $\delta$.
        Videos falling within $[\delta, \Delta]$ are annotated by humans.
        We then define the \textit{human labor reduction} as the highest possible proportion of videos labeled by MLLM, under the condition that the overall annotation quality remains no worse than that of purely manual labeling.
        Different configurations of $\Delta$ and $\delta$ yield different combinations of overall annotation quality and the proportion of videos labeled by MLLM.
        We conduct experiments on the SSC issue and performed a grid search for $\Delta$ and $\delta$.
        Each threshold configuration is plotted in Figure \ref{fig:cost}, where the x-axis represents the proportion of MLLM-labeled videos and the y-axis represents the overall annotation quality.
        We also include a reference line showing the quality level of manual annotation ($\text{F1}=0.730$).
        The upper-left region of the plot forms the Pareto frontier.
        Among all configurations that surpass the human annotation quality, the one with the highest proportion of MLLM-labeled videos achieves 0.382.
        This indicates that up to 38.2\% of human labor can be saved theoretically for the SSC issue.

    \subsection{Comparison with Content Governance Models}
    \label{app:experiment_compare_governance}

        \begin{table}[t]
            \centering
            \setlength{\tabcolsep}{7pt}
            \small
            \begin{tabular}{c|ccc}
                \toprule
                \textbf{Model} & \textbf{SSC} & \textbf{CB} & \textbf{IPC} \\
                \midrule
                \midrule
                Content governance model & 30.5 & 57.4 & 43.7 \\
                \midrule
                Discriminative MLLM labeler & \textbf{33.2} & \textbf{65.3} & \textbf{47.4} \\
                \bottomrule
            \end{tabular}
            \caption{Comparison between content governance models and discriminative MLLM labeler. The evaluation metric is R@P50 (Recall at 50\% Precision) (in \%). The highest value in each column is shown in \textbf{bold}.}
            \label{tab:online}
        \end{table}
        
        The governance models are lightweight vision models that we have deployed online.
        Each issue corresponds to a specific governance model, which is used to detect whether the input video contains that issue.
        The base models for the governance models include VLMo, LLaVA, and Qwen.
        We fine-tune these base models using human-labeled binary video data.
                
        The result of comparison between online content governance models and our discriminative MLLM labeler is presented in Table \ref{tab:online}.
        Experimental results demonstrate that our discriminative MLLM labeler significantly outperforms the online content governance model across multiple issues.
        Specifically, it achieves relative improvements R@P50 of 8.9\%, 13.8\%, and 8.5\% on SSC, CB, and IPC, respectively.
        
        Although the generative MLLM labeler cannot be directly compared with the online content governance model due to differences in evaluation metrics, the comparison between the two types of labelers in Table \ref{tab:main} suggests that the generative MLLM labeler would yield even greater improvements.
        This substantial performance gap indicates that the content governance model can benefit from the labels generated by the labeling models, thereby continually enhancing its capabilities.

\onecolumn
\section{Prompt Design for Agentic Labeling}
    \label{app:prompt}
    
    The overall prompt used for the agent is as follows:
    \begin{lstlisting}
You are a guideline detection agent that works through systematic analysis.
WORKFLOW:

STEP 1: Video Content Analysis
- Get video features
- If details in the video are unclear, use zooming_in_video_frames to magnify or inspect specific regions.

STEP 2: Guideline Reading and Analysis
- MUST use read_guideline to read each relevant guideline before analysis

STEP 3: Detailed Guideline-by-Guideline Detection
- For each of the guidelines, systematically evaluate:
  * Does the video content match the violation criteria?
  * Are there specific indicators present?
  * Do any exceptions apply?
  * You could even try to split the guideline into several sub-guidelines and detect them one by one, if you think the guideline is too long.
  * If you are uncertain, use referencing_precedents to check similar annotated examples, or use rethinking_annotation_guidelines to review edge-case criteria.
  * **Conclusion**: VIOLATES / DOES NOT VIOLATE

STEP 4: Final Determination
- Summarize findings for all guidelines
- Make final judgment with clear reasoning
- Format: <Tier1_result>Normal/Non-Deceptive Bait/Deceptive Bait/No Video Features</Tier1_result> <Tier2_result>Emotional Bait/Wait Until the End/...</Tier2_result> 

CRITICAL RULES:
- Always use read_guideline for reading raw guideline text
- Always use detect_guideline tool for guideline detection
- You may also use referencing_precedents, rethinking_annotation_guidelines, and zooming_in_video_frames when needed to ensure accuracy and consistency
    \end{lstlisting}

    \vspace{0.5em}
    
    The prompt for Rethinking tool is as follows:
    \begin{lstlisting}
You are a useful video content reviewer. 

## Detailed Policy

<detailed_policy>

## Task

Now based on the policy and the provided video information, determine if the policy applies. Respond with "yes" or "no".

**Does the video below violate the policy? (yes/no)**

If the answer is yes, also tell me the reason in the format like "Reason: blabla.<answer>Yes</answer>". If the answer is no, the format is like "Reason: blabla.<answer>No</answer>"

**Video Information:**
- Title: {title}
- Sticker Texts: {sticker_texts}
- Caption: {caption}
- ASR text: {asr} ASR text ends.
- OCR text: {ocr} OCR text ends.
- Frames: 
    \end{lstlisting}

    \vspace{0.5em}
    
    The prompt for Zoom-in tool is as follows:
    \begin{lstlisting}
You are a useful video content reviewer. 

Your task is to detect <region_of_interest> and where it is.

**Grid System Instructions:**
- Each frame is overlaid with a 5x5 grid, numbered from 000 to 024.
- Each grid cell represents a 72x128 pixel area.
- Grid cells are arranged left to right, top to bottom.
- Each grid cell is labeled with a 3-digit identifier, shown as **black text on a white background**, located at the **center** of the grid cell.
- Each frame also has its **frame index** (e.g., "frame: 1") displayed in the **top-left corner**, in **white background with red text**.
- You can see the faint grid lines and cell numbers in each frame.

Now, for each frame, tell me if the frames below contains any object of our interests. If there is, return the grid cell numbers that cover the target area in that frame. Remember that the results can be **different** for each frame! So please generate the cell ids for each frame independently.

Reminder:
Do not rely on previous answers to guess the next one.  

When providing grid cell numbers, use the format [cell1, cell2, cell3, ...]. Each cell number is a 3-digit identifier visible in the grid overlay. 

For each input frame, the format should be one of the two kinds below: 

<frame> {i} </frame>: <answer>No</answer>. / <answer>Yes</answer>, <grid_cells>[cell1, cell2, cell3, ...]</grid_cells>. 

Now starts the frames:
    \end{lstlisting}

\end{document}